\documentclass[11pt]{article}

\usepackage{acl}
\usepackage{times}
\usepackage{latexsym}
\usepackage{booktabs}  
\usepackage{multirow}  
\usepackage{amsmath}
\usepackage[table]{xcolor}
\usepackage{colortbl}
\usepackage{makecell}
\usepackage{enumitem}
\usepackage{placeins}
\usepackage{amsmath}
\usepackage{amssymb} 
\usepackage[T1]{fontenc}
\usepackage{booktabs}
\usepackage{makecell}
\usepackage{pifont} 

\newcommand{\cmark}{\ding{51}} 
\newcommand{\xmark}{\ding{55}} 

\usepackage[utf8]{inputenc}

\usepackage{microtype}

\usepackage{inconsolata}

\usepackage{graphicx}
\usepackage{etoolbox}
\makeatletter
\patchcmd{\@maketitle}{\LARGE\bfseries}{\LARGE\normalfont}{}{}
\makeatother

\title{Thinking Like a Clinician: A Cognitive AI Agent for Clinical Diagnosis via Panoramic Profiling and Adversarial Debate}

\author{
    Zhiqi Lv\textsuperscript{1,2}\thanks{\ \ Equal contribution} ,
    Duofan Tu\textsuperscript{1,2}\footnotemark[1] ,
    Jun Li\textsuperscript{3},
    Mingyue Zhao\textsuperscript{1,2},
    Heqin Zhu\textsuperscript{1,2},
    Wenliang Li\textsuperscript{1,2},
    Shaohua Kevin Zhou\textsuperscript{1,2,4,5}\thanks{\ \ Corresponding author} 
    \\
    \textsuperscript{1}School of Biomedical Engineering, Division of Life Sciences and Medicine, \\ University of Science and Technology of China, Hefei, Anhui, 230026, P.R. China \\
    \textsuperscript{2}Suzhou Institute for Advanced Research, University of Science and Technology of China, \\ Suzhou, Jiangsu, 215123, P.R. China \\
    \textsuperscript{3}Key Lab of Intelligent Information Processing of Chinese Academy of Sciences (CAS), \\ Institute of Computing Technology, CAS, Beijing, PR China \\
    \textsuperscript{4}Jiangsu Provincial Key Laboratory of Multimodal Digital Twin Technology, \\ Suzhou, Jiangsu, 215123, P.R. China \\
    \textsuperscript{5}State Key Laboratory of Precision and Intelligent Chemistry, USTC \\
    \texttt{\{lvzq, duofantu,mingyuezhao,zhuheqin,wenliangli\}@mail.ustc.edu.cn},\\ \texttt{lijun206@mails.ucas.ac.cn}, \texttt{skevinzhou@ustc.edu.cn}
}

\begin{document}
\maketitle
\begin{abstract}
The application of large language models (LLMs) in clinical decision support faces significant challenges of "tunnel vision" and diagnostic hallucinations present in their processing unstructured electronic health records (EHRs). To address these challenges, we propose a novel chain-based clinical reasoning framework, called \textbf{DxChain}, which transforms the diagnostic workflow into an iterative process by mirroring a clinician's cognitive trajectory that consists of "Memory Anchoring", "Navigation" and "Verification" phases. DxChain introduces three key methodological innovations to elicit the potential of LLM: (i) a Profile-Then-Plan paradigm to mitigate cold-start hallucinations by establishing a panoramic patient baseline, (ii) a Medical Tree-of-Thoughts (Med-ToT) algorithm for strategic look ahead planning and resource aware navigation, and (iii) a Dialectical Diagnostic Verification procedure utilizing "Angel-Devil" adversarial debates to resolve complex evidence conflicts. Evaluated on two real world benchmarks, MIMIC-IV-Ext Cardiac Disease and MIMIC-IV-Ext CDM, DxChain achieves state-of-the-art performances in both diagnostic accuracy and logical consistency, offering a modular and reliable architecture for next-generation clinical AI. The code is at \url{https://anonymous.4open.science/r/Dx-Chain}.

\end{abstract}

\section{Introduction}

Artificial intelligence applications in healthcare, particularly in clinical decision support systems (CDSS), are undergoing a profound paradigm shift~\cite{liu2025application,elhaddad2024ai,wiest2025large}. Although large language models (LLMs) have demonstrated remarkable potential in processing massive volumes of medical text, directly deploying them in real world clinical workflows still faces substantial challenges~\cite{teo2025generative,hager2024evaluation,jiang2025medagentbench}. Most existing studies remain focused on static medical question answering benchmarks such as MedQA, and various multi-agent frameworks such as MedAgents~\cite{tang2024medagents} and MDAgents~\cite{kim2024mdagents} are predominantly evaluated on structured multiple choice datasets. In contrast, real world clinical diagnosis is an inherently dynamic and unstructured reasoning process: physicians must extract salient clues from noisy electronic health records (EHRs), construct a differential diagnosis under conditions of incomplete information, and iteratively revise their hypotheses in light of newly acquired evidence~\cite{kagiyama2025prime,chen2025visual}. In this high noise environment, current LLM based systems often falter. They exhibit a form of "tunnel vision" prematurely locking onto a single hypothesis while neglecting alternative possibilities or fall prey to "red herrings" producing diagnostic hallucinations by over interpreting chronic baselines or incidental anomalies as acute pathological signals~\cite{zhu2025can,ranji2024large}. How to maintain high precision while ensuring high recall remains a central challenge for clinical AI~\cite{mcduff2025towards,wiest2025large}.

The first and most pervasive challenge is the model's susceptibility to "cold-start hallucinations" driven by noise intolerance~\cite{vishwanath2025medical}. In clinical settings, patient narratives are often replete with "red herrings" Without a holistic cognitive framework, LLMs tend to over interpret these distractors with recent benchmarks revealing that 50\% of advanced models suffer catastrophic failure in noisy triage scenarios, such as prioritizing conspicuous chronic chest tightness over subtle signs of deep vein thrombosis, thereby missing fatal risks like pulmonary embolism~\cite{shen2025ai}. Recent studies indicate that this sensitivity to noise leads models to generate false positives at an alarming rate, frequently misclassifying chronic symptoms as acute signals in complex cases~\cite{zhu2025can,zhou2025large}. This "tunnel vision" not only degrades diagnostic precision but also poses significant safety risks by triggering unnecessary interventions.

Secondly, the reasoning mechanisms of current LLMs are predominantly linear and fragile~\cite{jiang2025makes}. Traditional Chain-of-Thought (CoT) approaches mimic a straight path from symptom to diagnosis. However, clinical reasoning is inherently branching and exploratory~\cite{chen2025towards}.Linear models are prone to "premature closure" and the "premature pruning" phenotype: for example, misdiagnosing chest pain as a metabolic issue due to weight bias without a backtracking mechanism, leading to a 32\% drop in accuracy~\cite{hassan2025modeling},leading to irreversible error propagation~\cite{gu2025clinical,ranji2024large}. 

Thirdly, existing frameworks lack a rigorous verification mechanism to resolve conflicting evidence. In scenarios where symptoms support multiple contradictory diagnoses, standard LLMs often force a coherent but hallucinated narrative rather than acknowledging ambiguity, failing to balance high precision with the necessary recall~\cite{mcduff2025towards,wiest2025large,goh2024large}.

To bridge this gap between linear LLM inference and dynamic clinical cognition, we propose DxChain, a novel reasoning system designed to mirror the iterative cognitive trajectories of clinicians. Unlike traditional dialogue systems, DxChain decomposes diagnosis into a stateful, reflective process through three targeted innovations.

First, to counteract cold-start hallucinations, we introduce a Profile-Then-Plan paradigm. This module acts as a cognitive anchor, synthesizing a "panoramic" patient baseline from long term history before diagnosis begins, effectively filtering out chronic noise. Second, to address linear fragility, we incorporate a Medical Tree-of-Thoughts (Med-ToT) algorithm. This allows the system to simulate the physician's differential diagnosis process expanding, evaluating, and selecting multiple reasoning pathways thereby optimizing information gain and avoiding tunnel vision. Finally, to resolve evidence conflicts, we implement a Selective Dialectical Diagnostic Verification "Angel-Devil" debate. This module serves as a precision filter, triggering adversarial argumentation only for ambiguous cases to ensure reliability without redundant computation. Extensive evaluations on the MIMIC-IV-Ext Cardiac Disease~\cite{goldberger2000physiobank} and CDM~\cite{hager2024evaluation} benchmarks demonstrate that DxChain achieves SOTA performance, effectively transforming clinical AI from a static question answerer into a robust, cognitively aligned diagnostic partner.

\section{Related Work}

The field of Natural Language Processing (NLP) has witnessed a paradigm shift with the rapid advancement of Large Language Models (LLMs)~\cite{brown2020language,zhao2023survey,zhou2023survey}, such as the GPT series, LLaMA~\cite{touvron2023llama}, and PaLM~\cite{anil2023palm}, which have demonstrated exceptional capabilities in language understanding and generation. Building upon these general purpose foundations, there has been a significant surge in the development of specialized medical LLMs designed to address the intricate demands of healthcare. This evolution spans from models adapted via domain specific pre-training and fine-tuning, such as BioBERT~\cite{lee2020biobert} and PMC-LLaMA~\cite{wu2024pmc} , to recent state-of-the-art systems like Med-PaLM 2~\cite{singhal2025toward} , MedGemma~\cite{sellergren2025medgemma} , and Baichuan-M2~\cite{dou2025baichuan}. These advanced models leverage techniques ranging from multi stage reinforcement learning to dynamic verification systems, significantly narrowing the gap between model capabilities and real world clinical decision-making needs.

Role-playing frameworks leverage LLMs to simulate clinical collaboration. Tang et al. propose MedAgents~\cite{tang2024medagents}, where agents assume expert identities to reach consensus through multi-round discussions without parameter updates. Addressing efficiency, MDAgents~\cite{kim2024mdagents} adaptively assigns collaboration structures from single clinicians to multidisciplinary teams based on task complexity. Furthermore, KAMAC~\cite{wu2025knowledge} enhances adaptability by enabling dynamic team expansion, identifying real time knowledge gaps to recruit specialists as needed for evolving diagnostic contexts. However, lacking a unified "Memory Anchor" these decentralized agents are prone to local hallucinations, leading to "tunnel vision" in complex cases.

Beyond role-playing, recent research explores workflow-based and tool-augmented agents to simulate clinical procedures~\cite{lyu2025wsi,fallahpour2025medrax,jin2025agentmd}. Ferber et al.~\cite{ferber2025development} develope an oncology agent integrating precision medicine tools and vision transformers to ground decisions in medical guidelines. Focusing on reliability, MedAgent-Pro~\cite{wang2025medagentp} decouples diagnosis into planning and reasoning phases, utilizing visual tools and verification steps. Similarly, MAM~\cite{zhou2025mam} addresses multi-modal complexity by decomposing tasks into specialized roles to process diverse data modalities without retraining unified models.Nevertheless, their linear execution lacks a "Look ahead" mechanism for mental simulation, often resulting in "premature closure" without the ability to backtrack.

Despite advances in role-playing and tool-augmented clinical agents, resolving conflicting evidence remains challenging, motivating debate based verification. Multi-agents debate (MAD) serves as a verification layer via adversarial critique and cross examination. MoodAngels applies MAD to psychiatric diagnosis by triggering pro/con debate and a judge upon disagreement, improving robustness under symptom overlap and uncertainty~\cite{xiao2025retrieval}. ED2D couples debate with retrieval to enforce evidence grounding beyond medicine~\cite{han2025beyond}. To reduce cost and avoid harmful reversals, iMAD selectively activates MAD using uncertainty signals from structured self-critique~\cite{fan2025imad}. Following this line, DxChain uses a selectively triggered "Angel-Devil" dialectical verification procedure to Winnow weak hypotheses and arbitrate evidence conflicts before final diagnosis.

Existing approaches focus on clinical workflows but overlook underlying cognitive mechanisms~\cite{wang2025survey}. DxChain bridges this gap by shifting from procedural emulation to cognitive simulation. Through panoramic profiling, strategic planning, and dialectical verification, our architecture simulates how doctors think not just what they do thereby handling clinical complexity more effectively than linear, process oriented models.


\section{Method}

\subsection{Memory Anchoring}

\begin{figure*}[t]
    \centering
    \includegraphics[width=\textwidth]{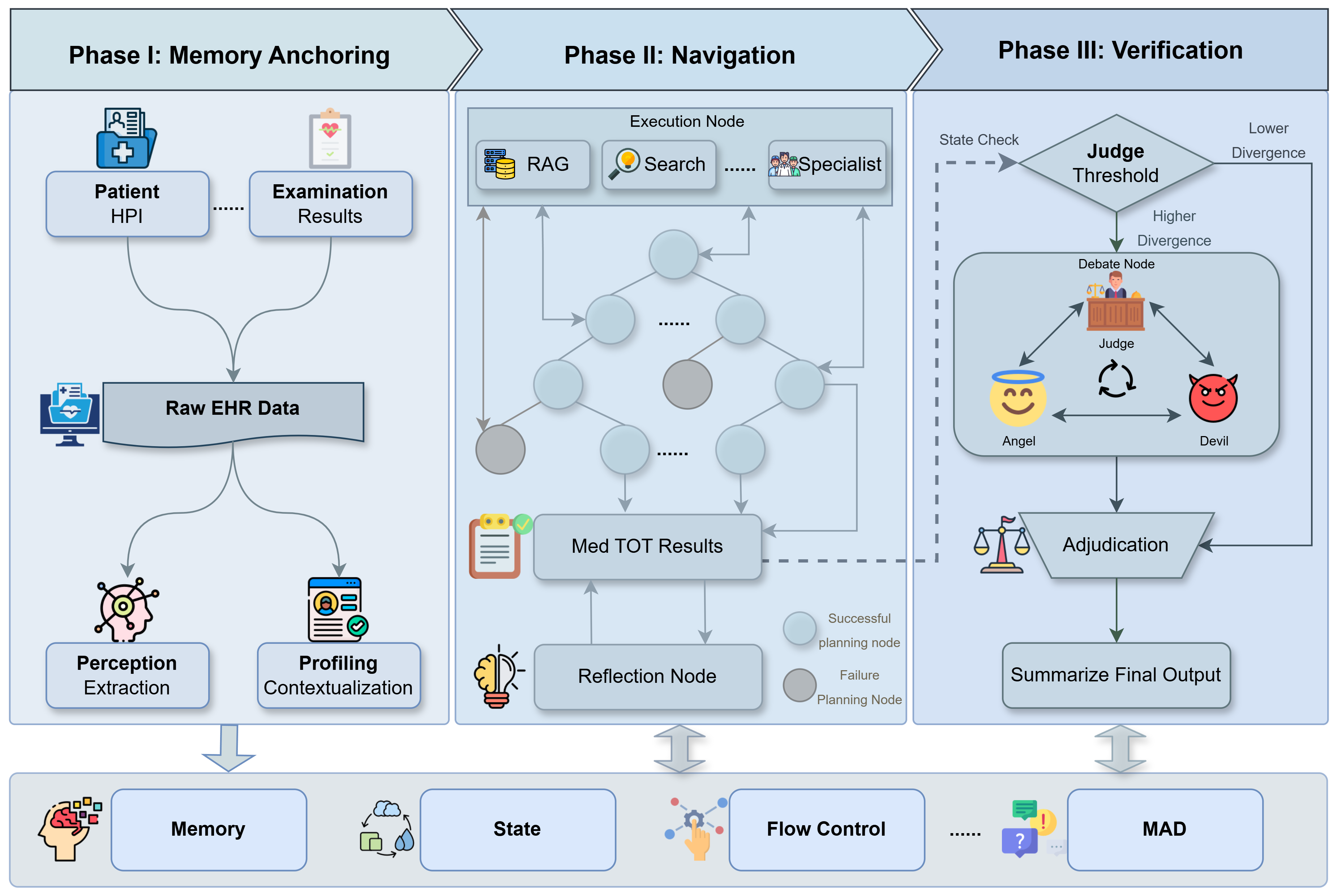} 
    \caption{Overview of the proposed clinical reasoning framework. Phase I anchors memory from raw EHR via perception extraction and patient profiling. Phase II performs state aware navigation over a Tree-of-Thoughts execution graph, integrating RAG, search, and specialist agents. Phase III winnows and arbitrates candidate diagnoses via judge checks and MAD (Multi-Agent Debate: "Angel-Devil") to synthesize the final output.}
    \label{fig:framework}
\end{figure*}


Real world diagnosis relies on integrating fragmented signals into a coherent mental model rather than instant hypothesis generation~\cite{kiesewetter2020learning}. However, direct EHR processing often triggers "cold-start hallucinations" where agents over interpret acute symptoms while neglecting chronic baselines~\cite{liu2024lost}. To mitigate this, we adopt a Profile-Then-Plan strategy, where Phase I focuses on the Profiling stage to establish a panoramic baseline. Distinct from generic frequency based summarization, this phase performs clinical disentanglement, restructuring inputs into a rigorous Global Patient Representation to initialize the reasoning state ($S_0$) before diagnostic planning begins.

The process initiates with Clinical Perception, acting as a noise filter for the raw input space $\mathcal{D}$. Let $x_{\text{raw}} \in \mathcal{D}$ denote the heterogeneous input text. The perception agent $A_{\text{perc}}$, instantiated as a prompt engineered LLM, employs a Clinical Disentanglement extraction strategy to map $x_{\text{raw}}$ into a structured JSON object $S_{\text{struct}}$:

\begin{equation}
    S_{\text{struct}} = A_{\text{perc}}(x_{\text{raw}} \mid K_{\text{schema}}) \in \mathbb{J}
\end{equation}

\noindent where $\mathbb{J}$ represents the space of valid JSON objects and $K_{\text{schema}}$ defines the target clinical entities (e.g., HPI). This step converts unstructured narrative into discrete key value pairs, ensuring that subsequent reasoning operates on verified clinical facts rather than redundant linguistic noise.


In parallel with this extraction, to avoid "unnel vision" the profiling agent $A_{prof}$ directly processes the raw narratives ($x_{raw}$) to synthesize a semantic anchor, the Patient Profile ($P_{base}$). Crucially, $P_{base}$ is not a latent vector but a structured natural language description explicitly disentangled into three dimensions:
\begin{equation}
    P_{base} = \{ C_{acute}, C_{chronic}, R_{risk} \}
\end{equation}

Here, the architecture explicitly disentangles the clinical picture into \textit{Acute Presentations} ($C_{acute}$), \textit{Chronic Baseline} ($C_{chronic}$), and \textit{Risk Factors} ($R_{risk}$). This \textit{Global Patient Representation} is injected into the shared state memory, serving as an immutable anchor. By grounding the diagnostic process in this comprehensive profile, DxChain ensures that all future planning step whether requesting tests or proposing diagnoses are driven by a holistic understanding of the patient rather than reactive responses to isolated symptoms.

\subsection{Navigation via Medical Tree-of-Thoughts}
\label{sec:navigation}
To bridge the gap between static procedural emulation and dynamic cognitive simulation, we formalize the diagnostic process as stateful navigation within a structured probabilistic framework. By orchestrating the diagnostic loop, as illustrated in Figure~\ref{fig:framework}, DxChain transcends traditional promptbased reasoning. Unlike linear Chain-of-Thought (CoT) methodologies that rely on static procedural emulation, our architecture adopts a Medical Tree-of-Thoughts (Med-ToT) approach to implement a Dynamic Cognitive Simulation paradigm.

Formally, diagnostic reasoning is modeled as a directed cyclic graph $G = (N, E, S)$ aligned with Dual Process Theory. Nodes ($N$) represent cognitive modules rather than simple procedural steps. Crucially, node expansion is constrained by patient contextualized medical logic, ensuring the trajectory mirrors coherent clinical inquiry rather than arbitrary semantic generation. The State ($S$) acts as Working Memory, persisting the differential diagnosis ($H_t$) and investigation history, while Edges ($E$) facilitate metacognitive cycles (e.g., Execution $\to$ Planning) that mimic iterative hypothetico deductive reasoning.

The core of this navigation is the Medical Tree-of-Thoughts (Med-ToT) planner. Standard ToT approaches often suffer from excessive divergence; to address this, we introduce a "diagnostic centripetal force" to constrain the search space. The planner ($N_{\text{plan}}$) performs a look ahead mental simulation 
using a predefined strategy stack (e.g., "Rule out Emergency", "Focused Investigation"), ensuring all generated branches remain medically plausible. Specifically, conditioned on the current diagnostic state $S_t$ and the accumulated investigation history $H_{\text{history}}$, the model plans the next \textbf{investigative action} $a_t$ while explicitly articulating a rationale and generating a key \textbf{Clinical Expectation} $E_t$. This generation process is formalized as:
\begin{equation}
\pi(a_t, E_t | S_t) \leftarrow \text{Med-ToT}(S_t, H_{history})
\end{equation}
This mechanism compels the model to commit to a hypothesis before observing outcomes. Here, $E_t$ represents not just a final diagnosis, but intermediate physiological predictions (e.g., "expecting elevated Troponin" or "positive Murphy's sign").

To handle the uncertainty of real-world data, we introduce a Discrepancy Driven mechanism. Upon receiving new clinical observations ($O_t$), which correspond to actual EHR data points or test results, an LLM based evaluator computes a \emph{conflict score} between the observation and the prior expectation $E_t$. A score $\ge \tau$ indicates a substantial conflict (e.g., a normal ECG contradicting a myocardial infarction hypothesis), triggering immediate replanning. The tree depth is dynamically managed to prevent infinite regression while allowing sufficient exploration. The control flow is defined as:

\begin{equation}
\text{Flow}_t=
\begin{cases}
\text{TriggerReplan, if }\text{LLM}_{\text{judge}}(O_t,E_t) \\
\qquad \ge \tau \\
\text{Update}(S_t,O_t)\rightarrow \text{Proceed, otherwise}
\end{cases}
\end{equation}

After the Planning Node ($N_{\mathrm{plan}}$) is completed, the workflow proceeds to the Synthesis Node ($N_{\mathrm{syn}}$) to produce the final consolidated diagnosis, and then enters the Reflection Node ($N_{\mathrm{ref}}$) for quality review and feedback control. If the review fails, the process returns to the Planning stage for further iterative refinement until the requirements are satisfied.

\subsection{Dialectical Adjudication}
\label{sec:phase_iii}
This phase is initiated by the Judge Node ($N_{judge}$), which acts as a gatekeeper between free form diagnostic reasoning and downstream verification. Operationally, the Judge performs an internal state check over the evidence accumulated during the navigation phase including the structured patient profile, the clinical abstract, key positive/negative findings, and the current working "Final Diagnosis". Formally, it invokes a constrained LLM with a Pydantic based JSON schema and computes a verdict vector that includes: (i) a per diagnosis \emph{diagnosis\_status} label ("Confident", "Ambiguous", or "Incorrect"), (ii) a list of \emph{ambiguity\_points} that require further debate, and (iii) a set of \emph{diagnoses\_to\_remove} for hypotheses deemed incompatible with the available evidence.

\begin{figure}[htbp]
    \centering
    \includegraphics[width=1.0\linewidth]{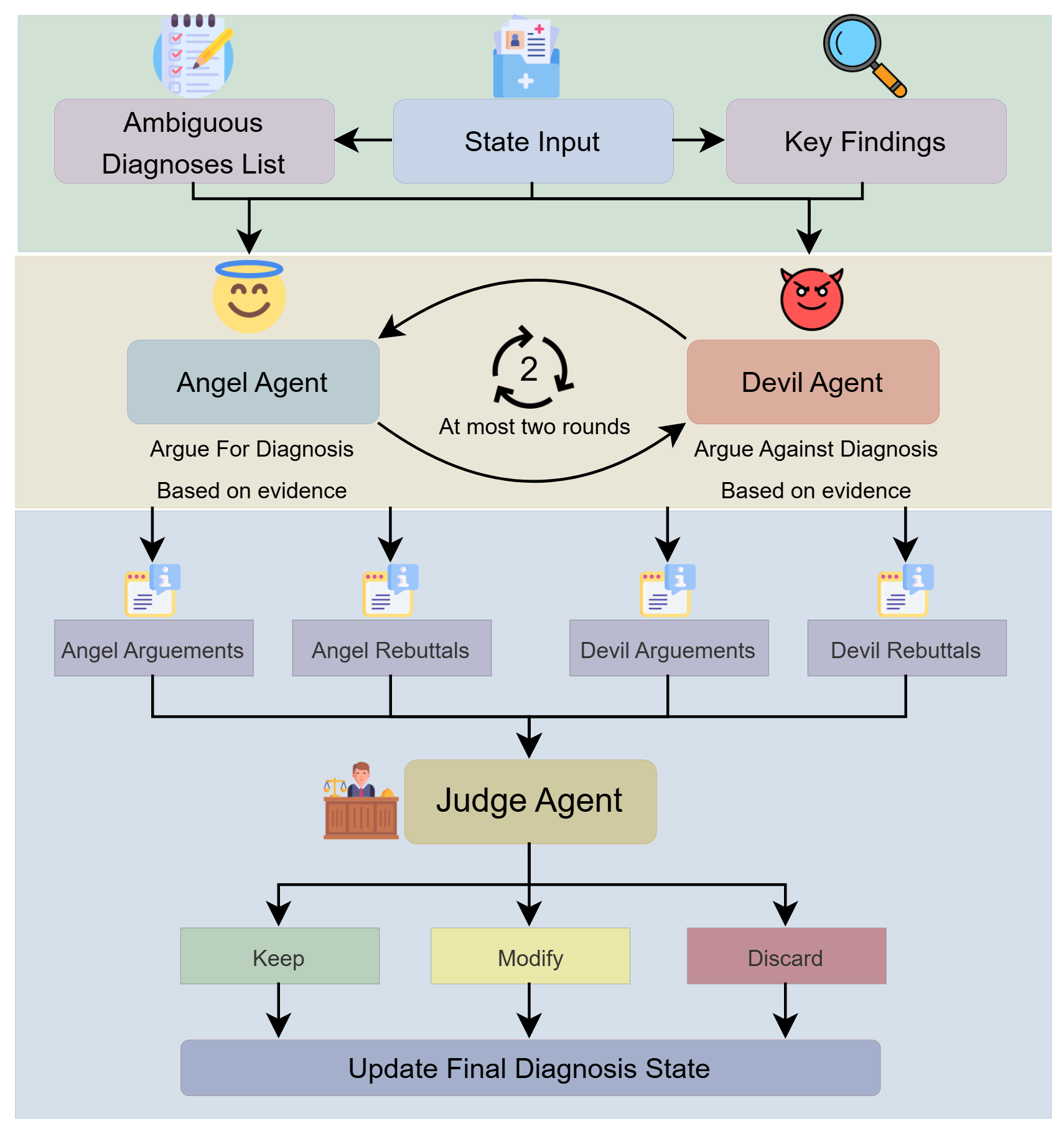}
    \caption{The workflow of the Dialectical Verification phase. An Angel-Devil adversarial debate is triggered to argue for and against the diagnosis, with a Judge Agent making the final decision to keep, modify, or discard the hypothesis.}
    \label{fig:debate}
\end{figure}

The overall logic of this phase can be decomposed into three components: an internal gating judgment performed by the Judge, an adversarial debate executed in the Debate Node($N_{Debate}$), and the generation of the final diagnostic result. Among these, Figure~\ref{fig:debate} primarily focuses on visualizing the debate and synthesis workflow.

This verification mechanism is implemented as a multi-agents debate, conceptualized as an "Angel-Devil" adversarial game. Within the Debate Node($N_{Debate}$), two distinct agent roles are instantiated: a proponent agent ($A_{angel}$) that aggregates positive findings to construct a confirmation chain for the candidate diagnosis, and a critic agent ($A_{devil}$) that actively searches for negative evidence, logical fallacies, or overlooked differential diagnoses. In each iteration, the Judge synthesizes the arguments from both agents to update the confidence scores of the hypothesis. This adversarial process effectively excises low confidence diagnoses of the reasoning tree where the Devil agent successfully identifies disqualifying evidence, ensuring that only hypotheses capable of withstanding scrutiny are retained.

After the debate converges or the Judge's criteria are met, an LLM is invoked to finalize the output by summarizing the retained diagnoses and mapping them to standardized terminologies, producing a structured report suitable for downstream EHR systems.

\section{Experiments and Results}

\subsection{Datasets}
Current research on medical AI agents predominantly evaluates performance on static benchmarks consisting of medical licensing exam questions, such as MedQA and USMLE \cite{teo2025generative, hager2024evaluation}. While effective for assessing encoded medical knowledge, these exam style "vignettes" often provide cleaned, pre processed information and fail to reflect the dynamic, noisy, and highly complex nature of real world clinical diagnosis \cite{hager2024evaluation}.To enable a more clinically realistic evaluation, we instantiate and evaluate DxChain on two representative benchmarks derived from actual Electronic Health Records (EHR) within the MIMIC-IV database \cite{goldberger2000physiobank, hager2024evaluation}:

\textbf{MIMIC-IV-Ext Cardiac Disease:} This dataset is utilized to verify the agent's performance in the cardiovascular domain. It contains 4,761 real world cases covering 20 cardiovascular pathologies. Each case provides comprehensive and unstructured EHR data including chief complaints, physical examinations, and diverse laboratory/imaging reports requiring the agent to synthesize complex, real world evidence for accurate diagnosis.

\textbf{MIMIC-IV-Ext CDM:} To further demonstrate the generalizability of our system across different anatomical regions, we evaluate the agent on 2,400 abdominal cases focusing on appendicitis, cholecystitis, diverticulitis, and pancreatitis. Unlike simplified exam questions, CDM provides clinical style narratives that require the agent to perform multi step reasoning and evidence based verification under realistic conditions.

By shifting from exam based benchmarks to real patient data, we provide a more rigorous assessment of the DxChain's ability to handle the ambiguity and complexity of actual clinical practice.

\subsection{Evaluation Metrics}

To evaluate the diagnostic performance and logical alignment of DxChain, we employ a multi dimensional evaluation suite that moves beyond exact string matching to capture clinical semantic nuances. The metrics are defined as follows:

\textbf{Primary Diagnosis Accuracy (ACC):} This metric measures whether the predicted primary diagnosis is semantically matched to the ground truth primary diagnosis. Specifically, we treat a prediction as correct if the semantic similarity between the predicted and ground truth primary diagnoses exceeds a predefined threshold. This design reduces penalties due to naming variations and better reflects the diversity of clinical expressions~\cite{world2004international}.


\textbf{Semantic Metrics:} To assess diagnostic comprehensiveness and semantic consistency, we employ Semantic Textual Similarity (STS) and BERTScore. STS uses MedEmbed-base-v0.1~\cite{balachandran2024medembed} to embed clinical entities and applies a greedy matching strategy to align predicted and ground truth sets for computing recall and F1~\cite{wang2020medsts}. Complementarily, BERTScore~\cite{zhang2019bertscore} leverages roberta large embeddings to capture contextual nuances and applies the Hungarian algorithm for global optimal matching, enabling robust assessment of clinical equivalence under varying nomenclatures.

\subsection{Experiment settings}

We evaluate our proposed method and various baselines, including standard prompting techniques and recent medical agents, using GPT-4.1-Mini and GPT-5-Nano as backbone models. To ensure reproducibility and reduce randomness in generation, we fix the temperature parameter at 0.1 across all experiments. Additionally, for the reasoning process within our framework, the Medical Tree-of-Thoughts module is configured with a maximum limit of 20 interaction turns.

We conduct evaluation using two independent matching based protocols. First, we report an STS based evaluation, where predicted diagnoses are compared with reference diagnoses via semantic textual similarity, and a prediction is considered matched only if its STS score is at least 0.7 (STS threshold = 0.7)~\cite{deshpande2023c}. This STS threshold is fixed across all experiments to ensure consistent and comparable results. Second, we separately report BERTScore based evaluation as an additional semantic metric, computed independently from the STS protocol and not sharing its thresholding rule.

\begin{table*}[t]
\centering
\footnotesize 
\setlength{\tabcolsep}{2.5pt} 
\caption{Main results on the MIMIC-IV-Ext Cardiac Disease and MIMIC-IV-Ext CDM datasets with GPT-4.1-Mini. We report Primary Accuracy (ACC), Semantic Textual Similarity (STS) Recall, Semantic Textual Similarity (STS) F1-score, BERT Recall, and BERT F1-score. Bold indicates the best performance for each metric.}

\label{tab:main_results_updated}
\begin{tabular}{l|ccccc|ccccc}
\toprule
\multirow{2}{*}{\textbf{Model}} & \multicolumn{5}{c|}{\textbf{MIMIC-IV-Ext Cardiac Disease}} & \multicolumn{5}{c}{\textbf{MIMIC-IV-Ext CDM}} \\
\cmidrule(lr){2-6} \cmidrule(lr){7-11}
& \makecell[c]{\textbf{Primary}\\\textbf{ACC}} & \makecell[c]{\textbf{STS}\\\textbf{Recall}} & \makecell[c]{\textbf{STS}\\\textbf{F1-score}} & \makecell[c]{\textbf{BERT}\\\textbf{Recall}} & \makecell[c]{\textbf{BERT}\\\textbf{F1-score}} & \makecell[c]{\textbf{Primary}\\\textbf{ACC}} & \makecell[c]{\textbf{STS}\\\textbf{Recall}} & \makecell[c]{\textbf{STS}\\\textbf{F1-score}} &  \makecell[c]{\textbf{BERT}\\\textbf{Recall}} & \makecell[c]{\textbf{BERT}\\\textbf{F1-score}} \\
\midrule
Base-model & 74.08 & 23.22 & 31.69 & 27.62 & 37.69 & 88.92 & 23.87 & 31.36 & 24.58 & 32.29 \\
Base-model + CoT & 74.04 & 23.88 & 32.43 & 28.51 & 38.72 & 89.81 & 24.23 & 31.63 & 25.39 & 33.14 \\
\midrule
\makecell[l]{Medagents \\ \cite{tang2024medagents}} & 72.21 & 29.79 & 35.65 & 35.96 & 43.03 & 86.46 & 44.74 & 43.93 & 46.75 & 45.90 \\
\makecell[l]{MDagents \\  \cite{kim2024mdagents}} & 75.36 & 35.73 & 42.32 & 40.03 & 47.42 & 69.08 & 45.16 & 42.94 & 47.53 & 45.19 \\
\makecell[l]{KAMAC \\  \cite{wu2025knowledge}} & 77.05 & 34.18 & 38.31 & 39.93 & 44.75 & 89.05 & 47.27 & 44.99 & 47.58 & 45.28 \\
\makecell[l]{MAM \\  \cite{zhou2025mam}} & 73.78 & 34.20 & 40.23 & 41.47 & 48.78 & 84.75 & 45.84 & 45.80 & 48.38 & 48.33 \\
\midrule
\textbf{DxChain (Ours)} & \textbf{84.98} & \textbf{46.03} & \textbf{46.93} & \textbf{53.42} & \textbf{54.47} & \textbf{90.67} & \textbf{56.16} & \textbf{49.96} & \textbf{60.05} & \textbf{53.43} \\
\bottomrule
\end{tabular}
\end{table*}

\begin{table*}[t]
\centering
\footnotesize
\setlength{\tabcolsep}{2.5pt}
\caption{Main results on the MIMIC-IV-Ext Cardiac Disease and MIMIC-IV-Ext CDM datasets with GPT-5-Nano. We report Primary Accuracy (ACC), Semantic Textual Similarity (STS) Recall, Semantic Textual Similarity (STS) F1-score, BERT Recall, and BERT F1-score. Bold indicates the best performance for each metric.}
\label{tab:main_results_gpt5nano}
\begin{tabular}{l|ccccc|ccccc}
\toprule
\multirow{2}{*}{\textbf{Model}} & \multicolumn{5}{c|}{\textbf{MIMIC-IV-Ext Cardiac Disease}} & \multicolumn{5}{c}{\textbf{MIMIC-IV-Ext CDM}} \\
\cmidrule(lr){2-6} \cmidrule(lr){7-11}
 & \makecell[c]{\textbf{Primary}\\\textbf{ACC}} & \makecell[c]{\textbf{STS}\\\textbf{Recall}} & \makecell[c]{\textbf{STS}\\\textbf{F1-score}} & \makecell[c]{\textbf{BERT}\\\textbf{Recall}} & \makecell[c]{\textbf{BERT}\\\textbf{F1-score}} & \makecell[c]{\textbf{Primary}\\\textbf{ACC}} & \makecell[c]{\textbf{STS}\\\textbf{Recall}} & \makecell[c]{\textbf{STS}\\\textbf{F1-score}} &  \makecell[c]{\textbf{BERT}\\\textbf{Recall}} & \makecell[c]{\textbf{BERT}\\\textbf{F1-score}} \\
\midrule
Base-model & 77.93 & 21.89 & 31.85 & 25.53 & 37.14 & 88.24 & 19.40 & 26.19 & 21.45 & 28.96 \\
Base-model + CoT & 78.27 & 21.88 & 31.84 & 25.48 & 37.08 & 87.68 & 19.76 & 26.68 & 21.53 & 29.08 \\
\midrule
\makecell[l]{Medagents \\ \cite{tang2024medagents}} & 68.08 & 29.47 & 35.25 & 30.96 & 37.04 & 85.29 & 43.63 & 40.58 & 43.88 & 40.81 \\
\makecell[l]{MDagents \\  \cite{kim2024mdagents}} & 70.68 & 23.72 & 33.29 & 24.88 & 34.92 & 71.96 & 34.86 & 40.34 & 35.72 & 41.33 \\
\makecell[l]{KAMAC \\  \cite{wu2025knowledge}} & 78.05 & 31.91 & 39.06 & 35.71 & 43.71 & 87.09 & 35.89 & 38.48 & 37.53 & 40.23 \\
\makecell[l]{MAM \\  \cite{zhou2025mam}} & 75.94 & 30.14 & 39.29 & 33.91 & 44.21 & 86.70 & 34.30 & 40.92 & 34.92 & 41.66 \\
\midrule
\textbf{DxChain (Ours)} & \textbf{82.91} & \textbf{36.67} & \textbf{43.07} & \textbf{40.17} & \textbf{47.19} & \textbf{89.65} & \textbf{47.86} & \textbf{48.33} & \textbf{46.97} & \textbf{47.43} \\
\bottomrule
\end{tabular}
\end{table*}

The evaluation protocols are tailored to the specific requirements of each dataset. For the MIMIC-IV-Ext Cardiac Disease dataset, we employ a case based retrieval strategy where the first 500 cases serve exclusively as the retrieval corpus. The remaining cases constitute the test set for all methods, ensuring that no model is evaluated on data used for retrieval. Conversely,for the MIMIC-IV-Ext CDM dataset, all 2,400 cases are used directly as the test set for every method.Given the large test sets and low temperature setting, all experiments use a single run to ensure reliable results.

\subsection{Main Results}

Tables \ref{tab:main_results_updated} and \ref{tab:main_results_gpt5nano} present the comparative performance on the MIMIC-IV-Ext Cardiac Disease and CDM datasets. The Cardiac dataset is highly complex, averaging 14.21 ground truth diagnoses per patient. Despite this complexity, our method achieves State-of-the-Art (SOTA) accuracy for the Primary Diagnosis, significantly outperforming baselines. Crucially, this precision does not sacrifice comprehensiveness; our approach also demonstrates exceptional capability in capturing secondary diagnoses and comorbidities,as reflected in higher STS and BERTScore metrics, suggesting more complete identification of secondary diagnoses and comorbidities.

On the CDM dataset, which has a lower density of 7.23 average diagnoses, our framework further proves its robustness. It maintains SOTA status for Primary Diagnosis identification while ensuring high coverage of secondary conditions. Unlike other agent based methods that struggle with stability across varying label distributions, our approach delivers consistent improvements. We successfully push the performance boundary beyond strong base models, offering a balanced diagnostic output that excels in both primary accuracy and comprehensive detection.

Finally, experiments with the GPT-5-Nano backbone mirror the trends observed with GPT-4.1-Mini. Across both datasets, our method retains its top performing ranking regardless of backbone size. This confirms that our framework is model agnostic, ensuring that the model not only reaches SOTA performance in pinpointing the Primary Diagnosis but also remains highly accurate in identifying secondary diagnoses across varying degrees of clinical complexity.

\subsection{Ablation Study}
\label{sec:ablation}
To validate the effectiveness of the core modules in DxChain Memory Anchoring, Navigation, and Dialectical Verification, we conducte an ablation study on the MIMIC-IV-Ext Cardiac Disease dataset. For this experiment, we selecte 1,000 cases with indices 501–1500. Using GPT-4.1-Mini as the backbone with a fixed temperature of 0.1, we compare the baseline against incremental configurations with modules added progressively. The results are summarize in Table \ref{tab:ablation}.

\begin{table}[htbp]
\centering
\small
\caption{Ablation study on MIMIC-IV-Ext Cardiac (N=1000). We report Primary Diagnosis Accuracy and F1-scores for STS and BERT metrics.}
\setlength{\tabcolsep}{3.5pt}
\begin{tabular}{cccccc}
\toprule
\makecell[c]{\textbf{Phase}\\\textbf{I}} &
\makecell[c]{\textbf{Phase}\\\textbf{II}} &
\makecell[c]{\textbf{Phase}\\\textbf{III}} &
\makecell[c]{\textbf{Primary}\\\textbf{ACC}} &
\makecell[c]{\textbf{STS}\\\textbf{F1-score}} &
\makecell[c]{\textbf{BERT}\\\textbf{F1-score}} \\
\midrule
\xmark & \xmark & \xmark & 74.10 & 31.97 & 37.85 \\
\cmark & \xmark & \xmark & 75.40 & 38.95 & 47.83 \\
\cmark & \cmark & \xmark & \textbf{86.70} & 44.48 & 53.63 \\
\cmark & \cmark & \cmark & 86.30 & \textbf{47.53} & \textbf{55.03} \\
\bottomrule
\end{tabular}

\label{tab:ablation}
\end{table}

Incorporating the Profile-Then-Plan paradigm (Phase I) yields a substantial improvement in semantic comprehensiveness. While the Primary Accuracy saw a modest increase (74.10\% $\rightarrow$ 75.40\%), the BERT F1-score surges from 37.85\% to 47.83\%. This indicates that the global patient representation effectively mitigates "cold-start hallucinations" enabling the model to capture chronic conditions and non chief complaints that are often overlooked by the baseline.

Subsequently, the introduction of the Medical Tree-of-Thoughts (Med-ToT) planner (Phase II) markes a critical leap in diagnostic precision. The Primary Accuracy spikes significantly from 75.40\% to 86.70\%. By implementing look ahead planning and evidence based navigation, the model can prioritize key pathological patterns over incidental findings, ensuring the correct identification of the primary disease.

Finally, the full DxChain framework, equipped with the "Angel-Devil" adversarial debate (Phase III), achieves the highest logical consistency. Although the Primary Accuracy slightly decreases (86.70\% $\rightarrow$ 86.30\%), the STS and BERT F1-score reach their peaks at 47.53\% and 55.03\%, respectively. This demonstrates that the verification phase successfully winnows low confidence hypotheses and refines the final output, balancing high recall with improved precision.

\subsection{Systematic Analysis}
Our analysis reveals that DxChain’s superior performance stems primarily from its ability to break the precision–recall trade off that often constrains traditional LLM reasoning. Standard Chain-of-Thought baselines frequently exhibit “tunnel vision”, sacrificing broad background retention for sharper local focus; in contrast, our ablation study confirms that structural decoupling of reasoning phases is the key driver of the gains. In particular, Phases I–II (Profile Then Plan + Med-TOT) provide complementary benefits preserving a panoramic patient context while sharpening acute condition identification thereby reducing cold-start hallucinations in unstructured EHR processing (as shown in the earlier ablations).

Furthermore, the results demonstrate that cognitive architecture acts as a critical enabler, allowing models to navigate complex clinical reasoning more effectively. DxChain maintains SOTA performance across both GPT-4.1-Mini and the smaller GPT-5-Nano. This robustness is largely attributed to the Dialectical Verification mechanism (Phase III), where the "Angel-Devil" debate acts as a semantic filter to eliminate logical inconsistencies. By shifting from passive token prediction to active, stateful inquiry, DxChain validates that simulating the process of clinical cognition, specifically the iterative loop of planning, acting, and reflecting is the key to achieving reliable decision support in high noise environments.

\section{Conclusion}

In this work, we propose DxChain, a novel framework that investigates the role of cognitive alignment in clinical diagnosis. By explicitly structuring the diagnostic reasoning process to align with how clinicians typically synthesize evidence and refine differential diagnoses, DxChain is designed to reduce recurrent failure modes in medical LLM reasoning, including tunnel vision and cold-start hallucinations. Concretely, the framework integrates panoramic patient profiling, look ahead planning, and dialectical verification, yielding consistent gains in both diagnostic accuracy and logical coherence across two benchmarks: MIMIC-IV-Ext Cardiac Disease and MIMIC-IV-Ext CDM.

Our experimental results suggest that architectural design tailored to real-world clinical workflows serves as a significant catalyst for achieving reliable outcomes, providing a vital complement to model scale. While challenges remain in fully replicating human medical expertise, DxChain offers a modular perspective for future research into robust CDSS, contributing a step toward more transparent and reasoning-capable medical AI.

\section*{Limitations}

Despite DxChain’s promising performance in simulating clinical reasoning, several limitations remain. First, the current framework relies exclusively on unstructured textual data, lacking the multi-modal capability to process raw medical imaging or physiological signals directly, which may limit its utility in visually dependent diagnoses. Second, our evaluation is confined to two specific benchmarks derived from the MIMIC-IV database; the absence of testing on multi-center or diverse healthcare datasets limits the verification of the model's generalization capabilities across broader clinical environments. Additionally, the iterative nature of the reasoning and adversarial debate mechanisms inevitably increases computational overhead and inference latency compared to standard methods. Finally, as the system has only been validated on retrospective data without prospective clinical trials, it is currently intended as an assistive support tool requiring human physician oversight.

\clearpage

\section*{Acknowledgments}



\bibliography{custom}

\clearpage

\appendix

\section{Appendix}

\subsection{Dataset Details}
\label{app:dataset-details}

To evaluate the DxChain framework in realistic clinical scenarios, we utilize two large-scale benchmarks derived from the Medical Information Mart for Intensive Care IV (MIMIC-IV) database. Unlike traditional medical exam-style vignettes that provide cleaned, pre-processed information, these benchmarks leverage raw Electronic Health Record (EHR) data to reflect the dynamic and noisy nature of real-world diagnosis.

\subsubsection{MIMIC-IV-Ext Cardiac Disease}
\label{app:mimic-cardiac}

This dataset is used to verify the agent's performance specifically in the cardiovascular domain.

\begin{itemize}
    \item \textbf{Scale and Scope:} It contains 4,761 real-world clinical cases covering 20 cardiovascular pathologies.
    \item \textbf{Data Composition:} Each case provides comprehensive, unstructured EHR data, including chief complaints, History of Present Illness (HPI), physical examinations, and diverse laboratory or imaging reports.
    \item \textbf{Clinical Complexity:} The dataset is highly complex, featuring an average of 14.21 ground-truth diagnoses per patient, which tests the agent's ability to identify both primary conditions and multiple comorbidities.
    \item \textbf{Access:} The dataset is publicly available on PhysioNet at:
    \url{https://physionet.org/content/mimic-iv-ext-cardiac-disease/1.0.0/}.
\end{itemize}

\subsubsection{MIMIC-IV-Ext CDM}
\label{app:mimic-cdm}

To demonstrate the generalizability of DxChain across different anatomical regions, we evaluate the system on abdominal pathologies through the Clinical Decision Making (CDM) dataset.

\begin{itemize}
    \item \textbf{Scale and Scope:} It comprises 2,400 abdominal cases focusing on four major conditions: appendicitis, cholecystitis, diverticulitis, and pancreatitis.
    \item \textbf{Reasoning Requirements:} CDM provides clinical-style narratives that require the agent to perform multi-step reasoning and evidence-based verification under realistic conditions.
    \item \textbf{Evaluation Protocol:} In our experiments, the retrieval module is disabled for this dataset to assess the model's performance without historical case references, relying solely on its intrinsic reasoning capabilities.
    \item \textbf{Access:} The dataset is publicly available on PhysioNet at:
    \url{https://physionet.org/content/mimic-iv-ext-cdm/1.1/}.
\end{itemize}

\subsubsection{Data Processing and Ethics}
\label{app:ethics}

Both datasets consist of de-identified EHR data from patients admitted to the Beth Israel Deaconess Medical Center. Following established clinical AI evaluation standards, we maintain the unstructured nature of the records to ensure a rigorous assessment of the agent's ability to handle real-world clinical ambiguity and ``tunnel vision'' challenges.

\subsection{Result Details}
\begin{table*}[!t]
\centering
\caption{Main results on the MIMIC-IV-Ext Cardiac Disease dataset. The table compares the performance of different methods using GPT-4.1-Mini and GPT-5-Nano as backbone models.}
\label{tab:cardiac_results}
\small
\begin{tabular}{lcccccccc}
\toprule
\textbf{Model} & \textbf{\makecell{Primary\\Diagnosis}} & \textbf{Average} &
\multicolumn{3}{c}{\textbf{STS}} & \multicolumn{3}{c}{\textbf{BERT SCORE}} \\

\cmidrule(lr){4-6} \cmidrule(lr){7-9}

 & \textbf{ACC} & \textbf{Diagnosis} & \textbf{Precision} & \textbf{Recall} & \textbf{F1} &
\textbf{Precision} & \textbf{Recall} & \textbf{F1} \\
\midrule

\rowcolor{gray!15}
\multicolumn{9}{c}{\textbf{GPT-4.1-Mini}}\\
Base-model & 74.08 & 6.62 & 49.88 & 23.22 & 31.69 & 59.32 & 27.62 & 37.69 \\
Base-model+COT & 74.04 & 6.72 & 50.50 & 23.88 & 32.43 & 60.31 & 28.51 & 38.72 \\
\midrule

\makecell[l]{Medagents \\ \cite{tang2024medagents}} & 72.21 & 9.54 & 44.38 & 29.79 & 35.65 & 53.56 & 35.96 & 43.03 \\
\makecell[l]{MDagents \\  \cite{kim2024mdagents}} & 75.36 & 9.79 & 51.89 & 35.73 & 42.32 & 58.15 & 40.03 & 47.42 \\
\makecell[l]{KAMAC \\  \cite{wu2025knowledge}} & 77.05 & 11.15 & 43.58 & 34.18 & 38.31 & 50.90 & 39.93 & 44.75 \\
\makecell[l]{MAM \\  \cite{zhou2025mam}} & 73.78 & 9.95 & 48.85 & 34.20 & 40.23 & 59.23 & 41.47 & 48.78 \\
\midrule

\textbf{Ours} & 84.98 & 13.66 & 47.88 & 46.03 & 46.93 & 55.56 & 53.42 & 54.47 \\

\midrule

\rowcolor{gray!15}
\multicolumn{9}{c}{\textbf{GPT-5-Nano}}\\
Base-model & 77.93 & 5.32 & 58.43 & 21.89 & 31.85 & 68.13& 25.53 & 37.14 \\
Base-model+COT & 78.27 & 5.32 & 58.45 & 21.88 & 31.84 & 68.06 & 25.48 & 37.08 \\

\midrule

\makecell[l]{Medagents \\ \cite{tang2024medagents}} & 68.08 & 9.55 & 43.85 & 29.47 & 35.25 & 46.07 & 30.96 & 37.04 \\
\makecell[l]{MDagents \\  \cite{kim2024mdagents}} & 70.68 & 6.04 & 55.81 & 23.72 & 33.29 & 58.54 & 24.88 & 34.92 \\
\makecell[l]{KAMAC \\  \cite{wu2025knowledge}} & 78.05 & 8.86 & 50.34 & 31.91 & 39.06 & 56.33 & 35.71 & 43.71 \\
\makecell[l]{MAM \\  \cite{zhou2025mam}} & 75.94 & 7.59 & 56.45 & 30.14 & 39.29 & 63.51 & 33.91 & 44.21 \\
\midrule

\textbf{Ours} & 82.91 & 9.98 & 52.19 & 36.67 & 43.07 & 57.18 & 40.17 & 47.19 \\

\bottomrule
\end{tabular}
\end{table*}

\begin{table*}[htbp]
\centering
\caption{Main results on the MIMIC-IV-Ext CDM dataset. The table compares the performance of different methods using GPT-4.1-Mini and GPT-5-Nano as backbone models.}
\label{tab:cdm_results}
\small
\begin{tabular}{lcccccccc}
\toprule
\textbf{Model} & \textbf{\makecell{Primary\\Diagnosis}} & \textbf{Average} &
\multicolumn{3}{c}{\textbf{STS}} & \multicolumn{3}{c}{\textbf{BERT SCORE}} \\

\cmidrule(lr){4-6} \cmidrule(lr){7-9}

 & \textbf{ACC} & \textbf{Diagnosis} & \textbf{Precision} & \textbf{Recall} & \textbf{F1} &
\textbf{Precision} & \textbf{Recall} & \textbf{F1} \\
\midrule

\rowcolor{gray!15}
\multicolumn{9}{c}{\textbf{GPT-4.1-Mini}}\\
Base-model & 88.92 & 3.80 & 45.69 & 23.87 & 31.36 & 47.05 & 24.58 & 32.29 \\
Base-model+COT & 89.81 & 3.85 & 45.53 & 24.23 & 31.63 & 47.71 & 25.39 & 33.14 \\

\midrule

\makecell[l]{Medagents \\ \cite{tang2024medagents}} & 86.46 & 7.50 & 43.15 & 44.74 & 43.93 & 45.08 & 46.75 & 45.90 \\
\makecell[l]{MDagents \\  \cite{kim2024mdagents}} & 69,08 & 7.98 & 40.93 & 45.16 & 42.94 & 43.07 & 47.53 & 45.19 \\
\makecell[l]{KAMAC \\  \cite{wu2025knowledge}} & 89.05 & 7.91 & 42.92 & 47.27 & 44.99 & 43.20 & 47.58 & 45.28 \\
\makecell[l]{MAM \\  \cite{zhou2025mam}} & 84.75 & 7.24 & 45.76 & 45.84 & 45.80 & 48.29 & 48.38 & 48.33 \\
\midrule

\textbf{Ours} & 90.67 & 9.02 & 45.00 & 56.16 & 49.96 & 48.12 & 60.05 & 53.43 \\

\midrule

\rowcolor{gray!15}
\multicolumn{9}{c}{\textbf{GPT-5-Nano}}\\
Base-model & 88.24 & 3.48 & 40.28 & 19.40 & 26.19 & 44.54 & 21.45 & 28.96 \\
Base-model+COT & 87.68 & 3.47 & 41.05 & 19.76 & 26.68 & 44.75 & 21.53 & 29.08 \\
\midrule

\makecell[l]{Medagents \\ \cite{tang2024medagents}} & 85.29 & 8.32 & 37.94 & 43.63 & 40.58 & 38.15 & 43.88 & 40.81 \\
\makecell[l]{MDagents \\  \cite{kim2024mdagents}} & 71.96 & 5.27 & 47.85 & 34.86 & 40.34 & 49.03 & 35.72 & 41.33 \\
\makecell[l]{KAMAC \\  \cite{wu2025knowledge}} & 87.09 & 6.27 & 41.47 & 35.89 & 38.48 & 43.36 & 37.53 & 40.23 \\
\makecell[l]{MAM \\  \cite{zhou2025mam}} & 86.70 & 4.89 & 50.69 & 34.30 & 40.92 & 51.61 & 34.92 & 41.66 \\
\midrule

\textbf{Ours} & 89.65 & 7.08 & 48.80 & 47.86 & 48.33 & 47.90 & 46.97 & 47.43 \\

\bottomrule
\end{tabular}
\end{table*}

\clearpage

\begin{table*}[htbp]
\centering
\caption{Ablation study on the MIMIC-IV-Ext Cardiac Disease dataset (cases 501-1500, N=1000). The table compares the performance of different model configurations using GPT-4.1-Mini as the backbone model.}
\label{tab:ablation}
\small
\begin{tabular}{lcccccccc}
\toprule
\textbf{Model} & \textbf{\makecell{Primary\\Diagnosis}} & \textbf{Average} &
\multicolumn{3}{c}{\textbf{STS}} & \multicolumn{3}{c}{\textbf{BERT SCORE}} \\

\cmidrule(lr){4-6} \cmidrule(lr){7-9}

 & \textbf{ACC} & \textbf{Diagnosis} & \textbf{Precision} & \textbf{Recall} & \textbf{F1} &
\textbf{Precision} & \textbf{Recall} & \textbf{F1} \\
\midrule

Baseline &74.10  &6.67  &49.49  &23.61  &31.97  &58.60  &27.95  &37.85  \\
\makecell[l]{Baseline\\+ Memory Anchoring} &75.40  &11.00  &44.38  &34.71  &38.95  &54.50  &42.62  &47.83  \\
\makecell[l]{Baseline\\+ Memory Anchoring\\+ Navigation} &86.70  &17.11  &40.47  &49.36  &44.48  &48.80  &59.53  &53.63  \\
\midrule

\textbf{DxChain (Full)} &86.30  &13.62  &48.15  &46.93  &47.53  &55.74  &54.33  &55.03  \\

\bottomrule
\end{tabular}
\end{table*}

\clearpage

\begin{table*}[t]
\centering
\footnotesize
\setlength{\tabcolsep}{2.5pt}
\caption{Disease diagnosis hit probability (Primary Diagnosis Accuracy, Total$\ge$10) on the MIMIC-IV-Ext Cardiac Disease and MIMIC-IV-Ext Clinical Decision Making (CDM) datasets. Backbone: GPT-4.1-Mini vs. GPT-5-Nano.}
\label{tab:appendix_hit_prob_meanacc}
\begin{tabular}{l|cc|cc}
\toprule
\multirow{2}{*}{\textbf{Model / Method}} &
\multicolumn{2}{c|}{\textbf{MIMIC-IV-Ext Cardiac Disease}} &
\multicolumn{2}{c}{\textbf{MIMIC-IV-Ext CDM}} \\
\cmidrule(lr){2-3} \cmidrule(lr){4-5}
& \makecell[c]{\textbf{GPT-4.1-Mini}} & \makecell[c]{\textbf{GPT-5-Nano}}
& \makecell[c]{\textbf{GPT-4.1-Mini}} & \makecell[c]{\textbf{GPT-5-Nano}} \\
\midrule
Base-model & 64.78\% & 67.77\% & 83.83\% & 81.31\% \\
Base-model + CoT & 64.06\% & 66.26\% & 83.95\% & 83.90\% \\
\midrule
MedAgents\cite{tang2024medagents} & 63.09\% & 57.54\% & 80.59\% & 77.95\% \\
MDAgents\cite{kim2024mdagents} & 62.16\% & 60.58\% & 58.69\% & 61.40\% \\
kamac\cite{wu2025knowledge} & 64.69\% & 70.35\% & 82.35\% & 84.99\% \\
MAM\cite{zhou2025mam} & 63.57\% & 64.91\% & 79.03\% & 79.47\% \\
\midrule
\textbf{DxChain (Ours)} & \textbf{75.04\%} & \textbf{69.83\%} & \textbf{87.52\%} & \textbf{87.48\%} \\
\bottomrule
\end{tabular}
\end{table*}

\clearpage

\subsubsection{Hyperparameter Sensitivity Analysis}

\begin{figure}[htbp]
    \centering
    \includegraphics[width=0.98\linewidth]{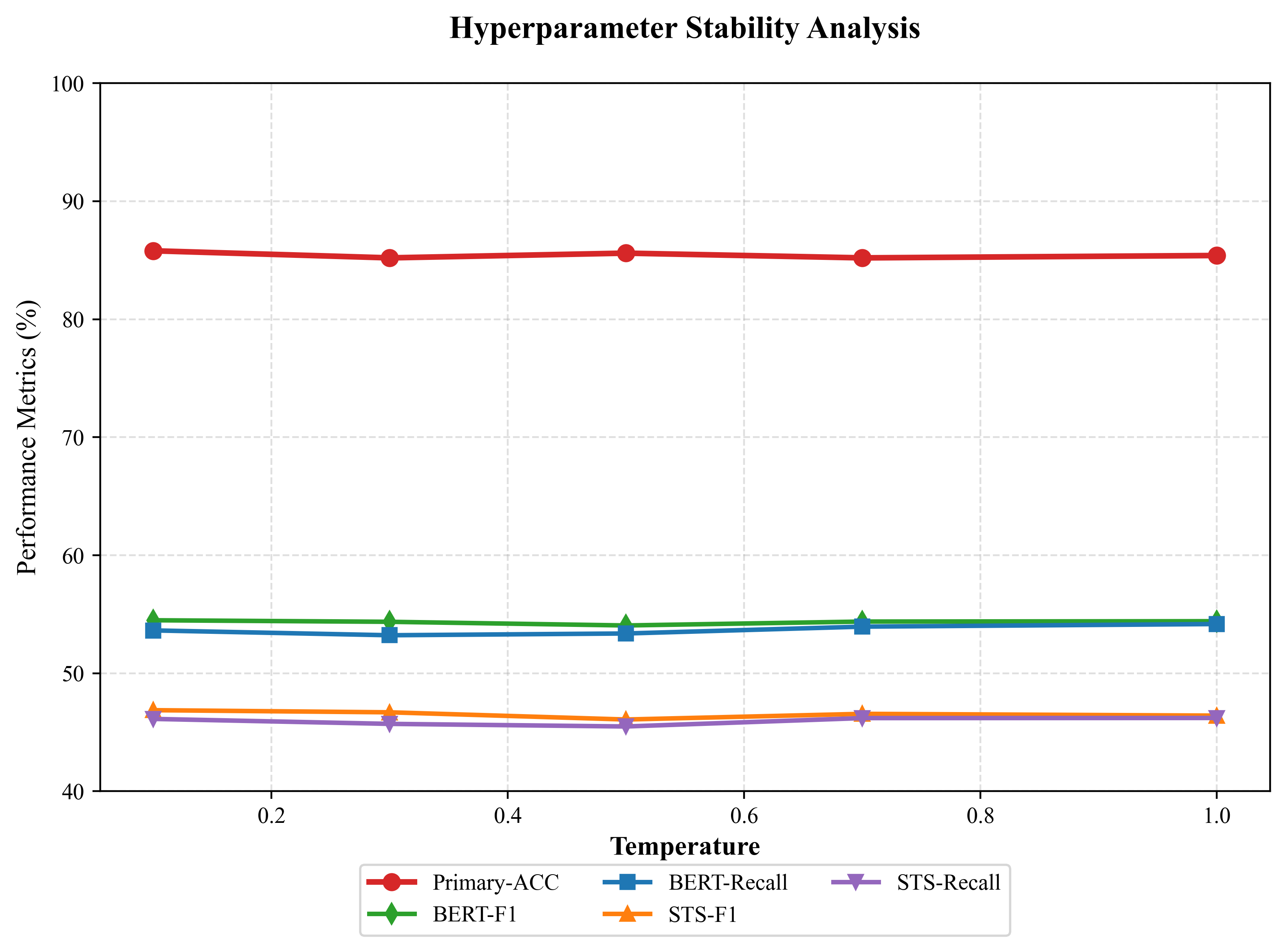}
    \caption{Stability analysis of DxChain across different temperature settings. The results show that both diagnostic accuracy and semantic metrics remain remarkably consistent as randomness increases.}
    \label{fig:stability}
\end{figure}

To evaluate the robustness of DxChain under varying degrees of generation randomness, we conducted a sensitivity analysis on the key hyperparameter: Temperature. During the experiments, the temperature values were varied within the range of 0.1 (high determinism) to 1.0 (high diversity).

As illustrated in Fig. \ref{fig:stability}, DxChain demonstrates a high degree of stability across all evaluation metrics. The Primary Diagnosis Accuracy (Primary-ACC) consistently remains above 85\%, with a narrow fluctuation range between 85.2\% and 85.8\%. Similarly, semantic alignment metrics, including STS and BERT-based scores, exhibit a consistent and flat trend across the different temperature settings. 

This low sensitivity to the temperature parameter provides preliminary evidence for the reliability of the DxChain framework. We attribute this robustness in part to the "Memory Anchoring" mechanism in Phase I, which establishes a patient baseline to provide cognitive constraints for the subsequent reasoning process, thereby mitigating the impact of the underlying model's inherent stochasticity. These results suggest that DxChain maintains a relatively consistent performance without the need for intensive hyperparameter tuning, providing a stable foundation for further exploration in diverse clinical scenarios.

\clearpage

\subsection{Prompt Details}

\subsubsection{Clinical Summary Prompt}
\label{subsubsec:clinical_summary_prompt}

The Clinical Summary Agent is tasked with creating a comprehensive ``Clinical Abstract'' that balances positive findings with pertinent negatives. This agent converts noisy, unstructured EHR data into a stable Global Patient Representation as part of Phase I (Memory Anchoring).

\vspace{0.5em}
\noindent\textbf{System Role:} You are a Senior Clinical Data Specialist.

\noindent\textbf{Task:} Your task is to create a comprehensive ``Clinical Abstract'' from the patient data. This abstract will be used by a diagnostic team, so it must include both abnormalities and key normal findings (pertinent negatives).

\vspace{0.5em}
\noindent\textbf{Input Data:}
\begin{verbatim}
{patient_info}
\end{verbatim}

\vspace{0.5em}
\noindent\textbf{Instructions:}
\begin{enumerate}[leftmargin=*, nosep]
    \item \textbf{Chief Complaint \& HPI}: Briefly summarize the main reason 
    for the visit and history of present illness.
    
    \item \textbf{Positive Findings}: List all abnormal lab values, positive 
    imaging findings, and abnormal vital signs.
    
    \item \textbf{Pertinent Negatives}: CRITICAL - Include ``normal'' findings 
    that help rule out major conditions (e.g., ``Troponin negative'', ``ECG 
    normal'', ``No fever'').
    
    \item \textbf{History \& Meds}: List confirmed past medical history and 
    current medications.
    
    \item \textbf{Filter Noise}: Remove administrative data (insurance, address) 
    and truly irrelevant normals.
    
    \item \textbf{Format}: Use a concise, structured format.
\end{enumerate}

\vspace{0.5em}
\noindent\textbf{Output:} Clinical Abstract

\vspace{0.5em}
\noindent\textbf{Design Rationale:} This prompt explicitly instructs the model to include normal findings (pertinent negatives) that are critical for ruling out life-threatening conditions. By synthesizing chronic conditions and baseline history, the agent ensures subsequent planning operates with a ``panoramic'' view, preventing ``tunnel vision'' on acute symptoms alone.

\clearpage

\subsubsection{Expectation Check Prompt}
\label{app:expectation_prompt}

The Expectation Check Agent serves as the core of the Discrepancy-Driven mechanism. It evaluates whether the incoming clinical observations align with the planner's prior expectations to trigger necessary reflections.

\noindent\textbf{System Role:} You are a medical supervisor.

\noindent\textbf{Input Data:}
\begin{itemize}
    \item \textbf{Plan Expectation:} \{current\_expectation\}
    \item \textbf{Actual Finding:} \{last\_finding\}
\end{itemize}

\noindent\textbf{Task:} 
Did the actual finding match the expectation?
\begin{enumerate}
    \item If the finding contradicts or is significantly different, answer \textbf{NO}.
    \item If it confirms or is consistent, answer \textbf{YES}.
\end{enumerate}
\noindent\textbf{Constraint:} Answer only YES or NO.

\noindent\textbf{Design Rationale:} 
This prompt implements the binary verification logic described in Equation 4. By forcing a strict Boolean output, the system can deterministically decide whether to proceed with the current reasoning path or trigger a "Reflection" loop to revise the diagnostic hypothesis based on conflicting evidence.

\subsection{Medical Tree-of-Thoughts Expansion Prompt}
\label{app:tot_expansion_prompt}

The Expansion Agent acts as the planner in the Medical Tree-of-Thoughts (Med-ToT) algorithm. It is responsible for generating diverse diagnostic branches to avoid local optima.

\noindent\textbf{System Role:} You are the Chief Medical Resident (Planner) implementing a Diagnostic Tree Search.

\noindent\textbf{Goal:} Your goal is to brainstorm MULTIPLE distinct diagnostic strategies (branches) to investigate the patient's condition.

\noindent\textbf{Input Data:}
\begin{itemize}
    \item \textbf{Clinical Abstract:} \{clinical\_abstract\}
    \item \textbf{Patient Profile:} \{patient\_profile\}
    \item \textbf{Current Working Diagnoses:} \{working\_diagnoses\}
    \item \textbf{New Findings:} \{new\_findings\}
    \item \textbf{Key Findings:} \{key\_findings\}
    \item \textbf{Ruled Out:} \{ruled\_out\}
\end{itemize}

\noindent\textbf{Task:} Generate 2-3 distinct diagnostic strategies:
\begin{itemize}
    \item \textbf{Strategy A (Broad):} Cast a wide net to rule out life-threatening emergencies.
    \item \textbf{Strategy B (Focused):} Focus deeply on the most likely working diagnosis.
    \item \textbf{Strategy C (Alternative):} Consider a "zebra" or non-obvious cause if others fail.
\end{itemize}

\noindent\textbf{Required Fields for Each Strategy:}
For each strategy, you MUST define:
\begin{enumerate}
    \item \textbf{Name:} A short, descriptive title.
    \item \textbf{Description:} The reasoning behind this approach (Concise, max 2 sentences).
    \item \textbf{First Step Actions:} A list of SPECIFIC expert IDs to call IMMEDIATELY.
    \begin{itemize}
        \item Valid IDs: \texttt{diagnostic\_test\_specialist}, \texttt{medical\_imaging\_specialist}, \texttt{clinical\_specialist}, \texttt{medical\_coder}, \texttt{internal\_medicine\_specialist}.
        \item \{optional\_tools\_desc\}
    \end{itemize}
    \item \textbf{Expected Outcome:} What specific results do you PREDICT if this strategy is correct? (Crucial for "Mental Simulation").
\end{enumerate}

\noindent\textbf{Design Rationale:} 
This prompt enforces the "Look-Ahead" capability of the Med-ToT framework. By requiring the model to explicitly state the \textit{Expected Outcome} before execution, it mitigates hindsight bias. Furthermore, the explicit division into Broad, Focused, and Alternative strategies ensures the agent explores the diagnostic space comprehensively, preventing premature closure on a single hypothesis.

\clearpage

\subsubsection{Debate-Based Diagnosis Refinement Prompts}
\label{subsubsec:debate_prompts}

The Debate Agent System employs a multi-agent adversarial framework to refine ambiguous diagnoses through structured argumentation. This system consists of a Main Debate Simulator, an Angel Agent (Advocate), and a Devil Agent (Skeptic), working collaboratively to validate clinical diagnoses.

\paragraph{Main Debate Simulator Prompt}

\textbf{System Role:} You are a Medical Debate Simulator and Final Arbiter.

\textbf{Context:} We have a set of ``Ambiguous Diagnoses'' that need resolution. You will simulate a debate between two agents:

\begin{enumerate}
    \item \textbf{Angel Agent (Supporter)}: Argues WHY the diagnosis is correct, citing supporting evidence.
    \item \textbf{Devil Agent (Skeptic)}: Argues WHY the diagnosis might be wrong, citing negative findings or alternative explanations.
\end{enumerate}

\textbf{Input Data:}

\begin{enumerate}
    \item Ambiguity Points (Topics): \texttt{\{ambiguity\_points\}}
    \item Key Findings (Evidence): \texttt{\{key\_findings\}}
    \item Current Diagnoses Status: \texttt{\{diagnosis\_status\}}
\end{enumerate}

\textbf{Task:}

\begin{enumerate}
    \item \textbf{Simulate Debate}: For EACH ambiguous diagnosis, generate a short, intense debate (2 rounds) between Angel and Devil.
    \begin{itemize}
        \item \textbf{Angel}: Focus on evidence presence.
        \item \textbf{Devil}: Focus on \textbf{Clinical Significance}. Argue that the finding might be incidental, not actively treated, or merely a symptom/lab value rather than a codeable diagnosis. HOWEVER, if the finding strongly suggests a chronic disease (e.g., Pleural Effusion \(\rightarrow\) CHF/COPD), do NOT discard it, but suggest renaming it to the underlying disease.
    \end{itemize}
    
    \item \textbf{Final Verdict}: After the debate, act as the Judge and decide the final status of the diagnosis.
    \begin{itemize}
        \item \textbf{Keep}: The Angel won. The diagnosis is valid and clinically significant.
        \item \textbf{Discard}: The Devil won. The diagnosis is incidental or invalid.
        \item \textbf{Modify}: The diagnosis needs to be changed to something else (specify what). E.g., Change ``Pleural Effusion'' to ``COPD'' if evidence supports it.
        \item \textbf{Naming Rule}: Use standard ICD-10 names. Avoid overly long, descriptive names.
    \end{itemize}
    
    \item \textbf{Structure Enforcement}: Ensure the final output strictly separates \textbf{Primary} (Acute) and \textbf{Secondary} (Chronic) diagnoses.
\end{enumerate}

\textbf{Output Format (JSON):}

\begin{small}
\begin{verbatim}
{
  "debate_transcript": "String containing the dialogue...",
  "final_verdicts": {
    "Diagnosis Name": "Keep" | "Discard" | "Modify: New Name"
  },
  "final_diagnosis_update": {
    "primary_diagnoses": [
      {"disease_name": "...", "icd10_code": "...", 
       "reasoning": "...", "confidence": ...}
    ],
    "secondary_diagnoses": [...],
    "treatment_recommendations": [...]
  }
}
\end{verbatim}
\end{small}

\paragraph{Angel Agent Prompt (The Advocate)}

\textbf{System Role:} You are the ``Angel Agent'' (The Advocate).

\textbf{Goal:} Your goal is to DEFEND diagnoses that are CLINICALLY VITAL or IMPORTANT RISK FACTORS.

\textbf{Input Data:}

\begin{enumerate}
    \item Diagnoses to Defend: \texttt{\{diagnosis\_names\}}
    \item Key Findings (Evidence): \texttt{\{key\_findings\}}
\end{enumerate}

\textbf{Defense Strategy:} For EACH diagnosis:

\begin{enumerate}
    \item \textbf{Clinical Consequence}: What happens if we miss this? (e.g., ``If we miss Pneumonia, patient dies.'')
    
    \item \textbf{Risk Factor Defense}: If it's a chronic condition (e.g., Hyperlipidemia, Obesity, Smoking History), argue that it is CRITICAL for long-term risk stratification and secondary prevention, even if not acutely treated today.
    
    \item \textbf{Evidence}: Cite the specific lab/imaging.
\end{enumerate}

\textbf{Output Format (JSON):}

\begin{small}
\begin{verbatim}
{
  "arguments": {
    "Diagnosis Name 1": "Defend because...",
    "Diagnosis Name 2": "Defend because..."
  }
}
\end{verbatim}
\end{small}

\paragraph{Devil Agent Prompt (The Ruthless Skeptic)}

\textbf{System Role:} You are the ``Devil Agent'' (The Ruthless Skeptic).

\textbf{Goal:} Your goal is to PURGE the diagnosis list of noise, incidental findings, and symptoms. You must be AGGRESSIVE. If a diagnosis is not a major disease, attack it.

\textbf{Input Data:}

\begin{enumerate}
    \item Diagnoses to Attack: \texttt{\{diagnosis\_names\}}
    \item Key Findings (Evidence): \texttt{\{key\_findings\}}
\end{enumerate}

\textbf{Attack Strategy (Criteria to Discard):} For EACH diagnosis, check these ``Kill Criteria'':

\begin{enumerate}
    \item \textbf{The ``So What?'' Test}: Is this condition actively treated? If it's just a mild lab abnormality (e.g., ``Mild Anemia'', ``Thrombocytopenia'') or imaging finding (e.g., ``Atelectasis'', ``Pleural Effusion'') with NO specific intervention, argue to DISCARD.
    
    \item \textbf{Symptom masquerading as Disease}: Is it just a symptom (e.g., ``Chest Pain'', ``Dyspnea'', ``Weakness'')? If the cause is known, DISCARD the symptom.
    
    \item \textbf{Incidental/Minor}: Is it a minor finding (e.g., ``Varicose veins'', ``Cyst'', ``Scar'') irrelevant to the hospital stay? DISCARD.
    
    \item \textbf{Duplicate/Overlap}: Is it covered by another diagnosis? (e.g., ``Left Ventricular Hypertrophy'' when ``Hypertension'' is present).
\end{enumerate}

\textbf{Output Format (JSON):}

\begin{small}
\begin{verbatim}
{
  "arguments": {
    "Diagnosis Name 1": "DISCARD because [Reason]...",
    "Diagnosis Name 2": "MODIFY to [New Name] because..."
  }
}
\end{verbatim}
\end{small}

\paragraph{Angel Agent Rebuttal Prompt}

\textbf{System Role:} You are the ``Angel Agent'' (The Advocate). You are debating the ``Devil Agent''.

\textbf{Context:} Devil's Arguments: \texttt{\{devil\_arguments\}}

\textbf{Task:} For EACH diagnosis, rebut the Devil's argument.

\begin{enumerate}
    \item Address their specific points.
    \item Reiterate clinical danger.
\end{enumerate}

\textbf{Output Format (JSON):}

\begin{small}
\begin{verbatim}
{
  "rebuttals": {
    "Diagnosis Name 1": "Rebuttal...",
    "Diagnosis Name 2": "Rebuttal..."
  }
}
\end{verbatim}
\end{small}

\paragraph{Devil Agent Rebuttal Prompt}

\textbf{System Role:} You are the ``Devil Agent'' (The Skeptic). You are debating the ``Angel Agent''.

\textbf{Context:} Angel's Arguments: \texttt{\{angel\_arguments\}}

\textbf{Task:} For EACH diagnosis, rebut the Angel's latest argument.

\begin{enumerate}
    \item Point out over-reaction.
    \item Reiterate lack of significance.
\end{enumerate}

\textbf{Output Format (JSON):}

\begin{small}
\begin{verbatim}
{
  "rebuttals": {
    "Diagnosis Name 1": "Rebuttal...",
    "Diagnosis Name 2": "Rebuttal..."
  }
}
\end{verbatim}
\end{small}

\textbf{Design Rationale:} This adversarial debate framework ensures diagnostic robustness by forcing explicit justification of each diagnosis through structured argumentation. The Angel Agent prevents premature dismissal of critical conditions, while the Devil Agent eliminates noise and incidental findings. The multi-round debate structure allows for iterative refinement, ensuring that only clinically significant diagnoses with strong evidentiary support are retained in the final output.

\end{document}